\documentclass[letterpaper, 10pt, conference]{ieeeconf}  

\IEEEoverridecommandlockouts                              

\newcommand{\etal}{\textit{et al}.}

\usepackage{graphicx}
\usepackage{amsmath}
\usepackage{amssymb}
\usepackage{booktabs}
\usepackage{lipsum} 
\usepackage[absolute,overlay]{textpos}
\usepackage{multirow}
\usepackage{url}
\usepackage{xcolor}
\usepackage{color}
\usepackage{pifont}
\usepackage{hyperref}
\usepackage{wrapfig}
\usepackage{multirow}
\usepackage{tabularx}
\usepackage{caption}
\usepackage{mathtools}
\usepackage{bm}
\usepackage{makecell}

\usepackage{colortbl}
\usepackage{dsfont}
\usepackage{bbm}
\hypersetup{pdfstartview=FitH,
            colorlinks=true,
            linkcolor=red,
            anchorcolor=blue,
            citecolor=black
            }
 
\definecolor{tablecolor}{HTML}{ccf2f5} 
\newcommand{\dd}[2]{$#1\scriptstyle{\pm#2}$}
\newcommand{\ddbf}[2]{$\mathbf{#1\scriptstyle{\pm#2}}$}
\newcommand{\ddone}[1]{$#1$}
\newcommand{\ddbfone}[1]{$\mathbf{#1}$}

\newcommand{\bepsilon}{{\boldsymbol{\epsilon}}}

\title{\LARGE \bf
\makebox[5pt][l]{\raisebox{-0.7ex}{\includegraphics[height=36pt]{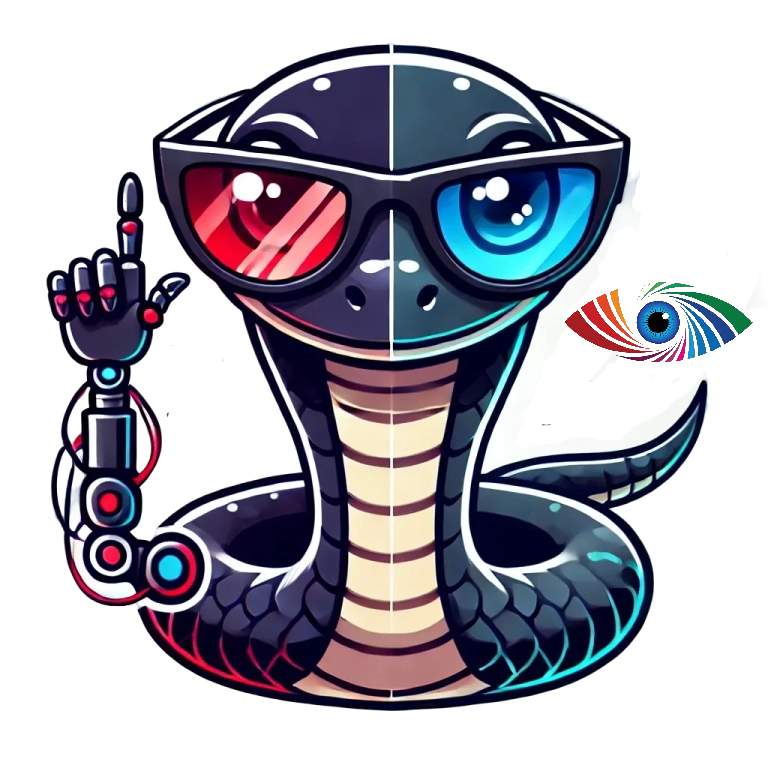}}}\hspace{28pt} Mamba Policy: Towards Efficient 3D Diffusion Policy with \\ Hybrid Selective State Models
}

\author{Jiahang Cao$^{1,3*}$, Qiang Zhang$^{1,3*}$, Jingkai Sun$^{1}$, Jiaxu Wang$^{1}$,
    Hao Cheng$^{1}$, Yulin Li$^{2}$,
    Jun Ma$^{2}$, \\ \quad Kun Wu$^{3}$, Zhiyuan Xu$^{3}$,
    Yecheng Shao$^{4}$, 
    Wen Zhao$^{3}$, 
    Gang Han$^{3}$,
    Yijie Guo$^{3}$, Renjing Xu$^{1\dagger}$
\thanks{$^\dagger$Corresponding author; $^*$Equal contribution.}
\thanks{$^{1}$Jiahang Cao, Qiang Zhang, Jingkai Sun, Jiaxu Wang, Hao Cheng, and Renjing Xu are with the Microelectronics Thrust, The Hong Kong University of Science and Technology (Guangzhou), China.
        {\tt\footnotesize jcao248@connect.hkust-gz.edu.cn;  renjingxu@ust.hk}}%
\thanks{$^{2}$  Yulin Li and Jun Ma are with the Division of Emerging Interdisciplinary
Areas, The Hong Kong University of Science and Technology, China.}
\thanks{$^{3}$Kun Wu, Zhiyuan Xu, Wen Zhao, Gang Han and Yijie Guo are with the Beijing Innovation Center of Humanoid Robotics, China.
}
\thanks{$^{4}$Yecheng Shao is with the Center for X-Mechanics, Zhejiang University, China.
}
}

\begin{document}

\maketitle


\begin{abstract}
Diffusion models have been widely employed in the field of 3D manipulation due to their efficient capability to learn distributions, allowing for precise prediction of action trajectories. However, diffusion models typically rely on large parameter UNet backbones as policy networks, which can be challenging to deploy on resource-constrained devices. Recently, the Mamba model has emerged as a promising solution for efficient modeling, offering low computational complexity and strong performance in sequence modeling.
In this work, we propose the Mamba Policy, a \textit{lighter but stronger} policy that reduces the parameter count by over \textbf{80\%} compared to the original policy network while achieving superior performance. Specifically, we introduce the XMamba Block, which effectively integrates input information with conditional features and leverages a combination of Mamba and Attention mechanisms for deep feature extraction. Extensive experiments demonstrate that the Mamba Policy excels on the Adroit, Dexart, and MetaWorld datasets, requiring significantly fewer computational resources. Additionally, we highlight the Mamba Policy's enhanced robustness in long-horizon scenarios compared to baseline methods and explore the performance of various Mamba variants within the Mamba Policy framework. Real-world experiments are also conducted to further validate its effectiveness. Our open-source project page can be found at
\url{https://sagecao1125.github.io/mamba_policy/}.

\end{abstract}

\section{Introduction}
\label{sec:intro}

Visuomotor policies, which involve the seamless integration of visual perception with motor control, are crucial for enabling robots to perform complex tasks based on visual inputs. 
Within this framework, imitation learning has emerged as an effective approach. By observing and mimicking human demonstrations, robots can learn a range of skills, such as dexterous hand control~\cite{arunachalam2023dexterous,wei2024wearable}, grasping~\cite{chi2023diffusion,ze20243d,yamane2023soft}, and locomotion~\cite{cheng2024extreme,tang2024humanmimic,seo2023deep}. Imitation learning simplifies the transfer of human expertise to robotic systems, making it a powerful tool for developing robots capable of performing intricate tasks.

\begin{figure}[tbp]
	\centering
	\includegraphics[width=1\linewidth]{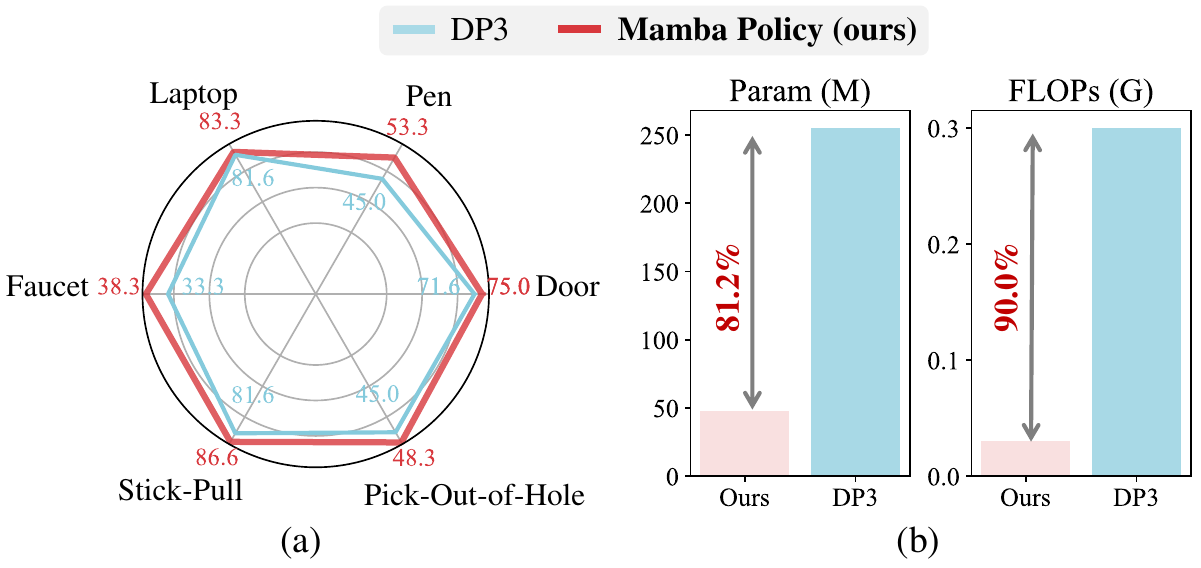}
 \caption{\textbf{Comparison with the SOTA baselines regarding accuracy and computational usage.} Our proposed Mamba Policy (a) achieves superior success rates and (b) offers up to $90\%$ computational savings in terms of floating point operations (FLOPs). }
	\label{fig:teaser}
	\vspace{-0.3cm}
\end{figure}
Among imitation learning methods, diffusion policy~\cite{chi2023diffusion} has recently attracted considerable attention. Its strength lies in its ability to effectively manage multimodal action distributions, a key factor in producing diverse and adaptable behaviors in complex environments. This makes diffusion policy particularly advantageous for robotic manipulation tasks, where flexibility and generalization are critical. Consequently, diffusion policies have been widely adopted across various robotic applications~\cite{ha2023scaling,kapelyukh2023dall,mishra2023generative,sridhar2024nomad}, consistently demonstrating impressive performance in diverse scenarios.

However, despite their success, diffusion policies typically depend on large backbone architectures. For instance, the 3D Diffusion Policy (DP3~\cite{ze20243d}) utilizes UNet models with over 200 million parameters. While these architectures excel at capturing intricate details and delivering high performance, their significant computational demands present challenges, especially in resource-constrained environments or when deployed on edge devices. Moreover, the ability to achieve efficient long-horizon prediction is also crucial in robot learning, as it enables more accurate decision-making and planning over extended periods, which is essential for complex tasks such as navigation~\cite{navarro2024sorts,song2022one} and manipulation~\cite{zhang2024universal,lee2021ikea}. These considerations highlight the need to develop efficient models that maintain the high performance of diffusion policies while significantly reducing computational overhead and improving long-term prediction capabilities.

One of the recent advancements, Mamba~\cite{gu2023mamba}, provides valuable insights into addressing these challenges with its selective state space model (SSM), which demonstrates low computational complexity while maintaining robust sequence modeling capabilities. This innovation has led to its increasing adoption in various robotics tasks\cite{liu2024robomamba,mustafa2024context,cao2024mamba}, as Mamba excels in trajectory modeling and effectively captures long-horizon dependencies, making it highly effective for complex motion planning and control.

In this work, we introduce Mamba Policy, a \textbf{lighter but stronger} policy that reduces the parameter count by over 80\% compared to the original policy network while delivering superior performance. We achieve this by integrating a hybrid state space model module with attention mechanisms~\cite{vaswani2017attention}, which we refer to as XMamba. To validate our approach, we conduct extensive experiments across multiple datasets, including Adroit~\cite{rajeswaran2017learning}, MetaWorld~\cite{yu2020meta}, and DexArt~\cite{bao2023dexart}. The results demonstrate that Mamba Policy not only significantly outperforms the 3D Diffusion Policy (DP3) in terms of performance but also drastically reduces GPU memory usage. As illustrated in Fig.~\ref{fig:teaser}, our method achieves better results with lower computational demands compared to DP3. Additionally, we investigate the impact of horizon length to assess Mamba Policy's capabilities under long-term conditions and explore the effects of various Mamba variants, offering a comprehensive analysis of the effectiveness of our proposed approach.

Our contributions can be summarized as follows:
\begin{itemize}
    \item We introduce Mamba Policy, a lighter yet stronger policy method based on a hybrid state space model integrated with attention mechanisms.
    \item Our extensive experiments demonstrate that Mamba Policy achieves up to a $\mathbf{5\%}$ improvement in success rate under a variety of manipulation datasets, while reducing the parameter count by $\mathbf{80\%}$.
    \item We explore the effects of different horizon lengths, confirming Mamba Policy's stability in long-term scenarios, and analyze the impact of various SSM variants on model performance.
\end{itemize}

\begin{figure*}[tbp]
	\centering
	\includegraphics[width=1\linewidth]{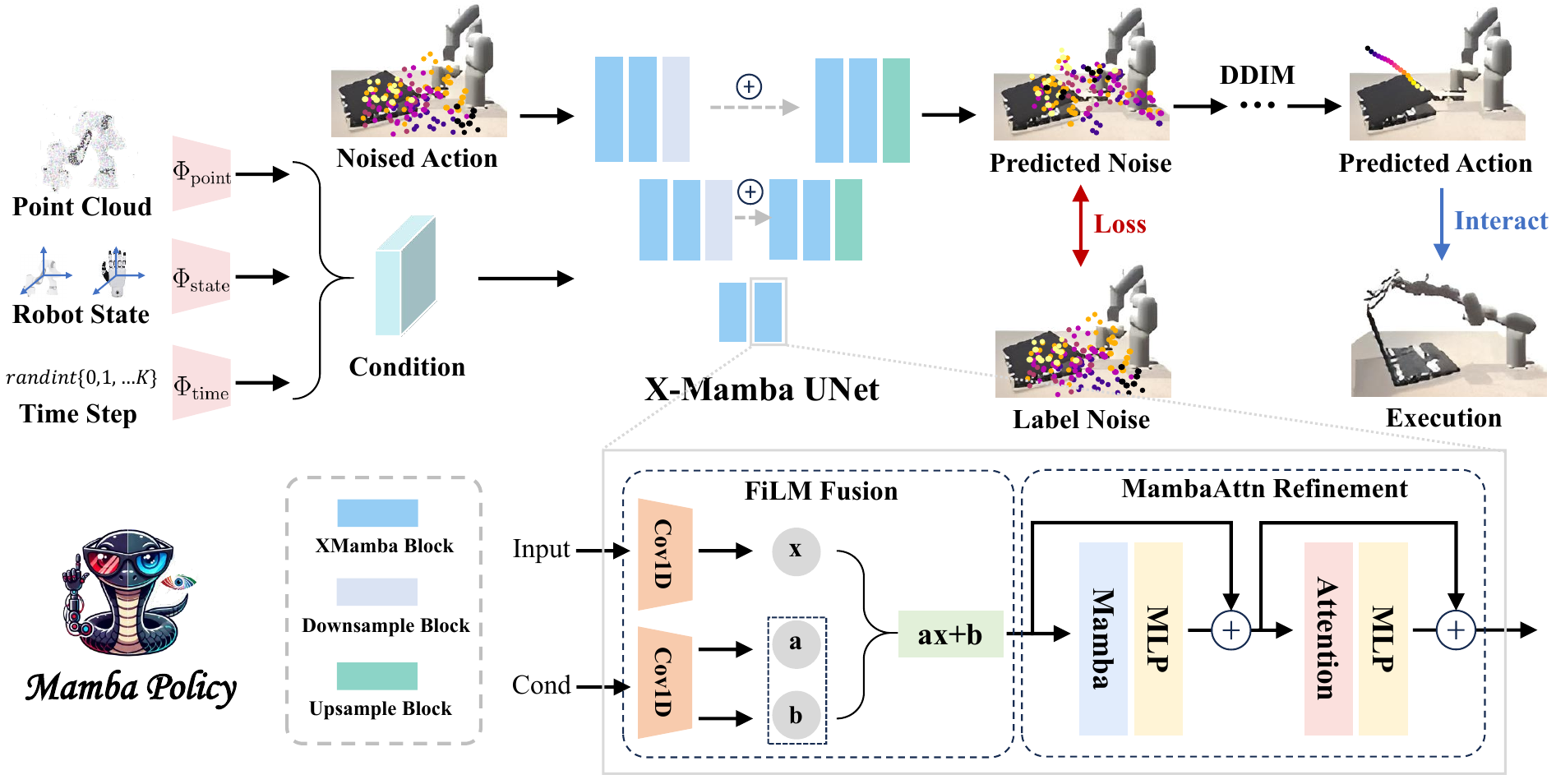}
 \caption{\textbf{Overview of Mamba Policy.} Our proposed model takes the noised action and the condition as inputs, the latter of which is composed of three parts: point cloud perception embedding, robot state embedding, and time embedding. Each of these components is processed through its respective encoder $\Phi_{\text{type}}$. The X-Mamba UNet is then employed to process these inputs and ultimately 
 return the predicted noise, with XMamba blocks serving as a key role. During training, the model is updated using MSE loss (Eq.~\ref{eq:loss}) with the label noise. For validation, the model leverages DDIM to reconstruct the original action, which is then used to interact with the environment and execute different tasks.}
	\label{fig:main}
\end{figure*}

\section{Related Work}
\label{sec:related_work}

\subsection{Diffusion Models in Robotic Manipulation}

Diffusion models, initially introduced in image generation, have recently gained significant traction in robotic manipulation due to their ability to generate complex and realistic trajectories by denoising random noise into desired actions or paths. Their flexibility and effectiveness have led to widespread adoption in various robotic tasks. Existing approaches can be broadly categorized into two main types based on the sensory information utilized.
The first category relies on 2D visual input, where diffusion models generate manipulation actions from RGB images. Chi~\etal~ introduced the Diffusion Policy (DP~\cite{chi2023diffusion}), where, instead of directly outputting an action, the policy infers the action-score gradient conditioned on 2D visual observations. Numerous works have been built on DP, applying it to a range of robotic tasks, such as grasping~\cite {mishra2023generative} and navigation~\cite{sridhar2024nomad}.
The second category leverages 3D visual information, such as point clouds, which offer a richer and more detailed representation of the environment. Recent 3D-based policies have achieved significant success in control tasks~\cite{gervet2023act3d,ke20243d,ze20243d,rana2023sayplan}.
In this work, our proposed Mamba Policy is based on 3D visual perception.

\subsection{State Space Models}
State Space Models (SSMs) represent a modern category of sequence models that draw inspiration from specific dynamic systems. 
To elucidate the process of modeling, we meticulously describe the architecture of Structured State Space models (S4) as follows:
\begin{align}
    h'(t) &= \mathbf{A}h(t) + \mathbf{B}x(t), \\
    y(t) &= \mathbf{C}h(t) + \mathbf{D}x(t),
\end{align}
with the quartet ($\mathbf{A}, \mathbf{B}, \mathbf{C}, \mathbf{D}$) steering the entire continuous framework. 
To handle discrete sequences, it's essential to convert S4 into a discretized variant:
\begin{align}
    h'(t) &= \overline{\mathbf{A}}h(t) + \overline{\mathbf{B}}x(t), \\
    y(t) &= \mathbf{C}h(t),
\end{align}
wherein S4 employs the Zero-Order Hold (ZOH) method for discretization, defined as $\overline{\mathbf{A}} = \exp (\Delta \mathbf{A})$ and $\overline{\mathbf{B}} = (\Delta \mathbf{A})^{-1}(\exp (\Delta \mathbf{A}) - \mathbf{I}) (\Delta \mathbf{B})$. Here, $\mathbf{D}$ is conceptualized as parameter-driven skip connections, hence simplified to 0 for brevity. Following the transformation from ($\Delta, \mathbf{A}, \mathbf{B}, \mathbf{C}, \mathbf{D}$) to ($\widetilde{\mathbf{A}}, \widetilde{\mathbf{B}}, \mathbf{C}$), the model can be executed through two perspectives: (a) a linear recurrence approach requiring only $O(1)$ complexity during inference and (b) a global convolution strategy that enables swift parallel processing during training sessions. This adaptive capability endows the SSM with a significant efficiency edge, especially when juxtaposed against traditional sequence models such as transformer~\cite{vaswani2017attention,dao2022flashattention,dao2023flashattention} and RWKV~\cite{peng2023rwkv} in NLP endeavors.

\subsection{Mamba and its Variants}

To elevate both selectivity and awareness of context, the introduction of Mamba~\cite{gu2023mamba} aims at tackling intricate sequential challenges. 
Mamba refines the S4 framework by modifying its core parameters to be time-dependent, as opposed to being time-independent. 
Such change significantly extends the utility of State Space Models (SSMs) to a wide array of fields, including, but not limited to, the realms of vision~\cite{hatamizadeh2024mambavision,zhu2024vision,liu2024vmamba,teng2024dim}, NLP~\cite{pioro2024moe,anthony2024blackmamba}, and healthcare~\cite{liu2024swin,xing2024segmamba}.

Many works have focused on improving the Mamba framework. Dao~\etal~\cite{dao2024transformers} developed a comprehensive framework of theoretical connections between SSMs and variants of attention, leading to the design of Mamba2 with state space duality, which achieves a 2-8$\times$ speedup compared to the original Mamba. Vision Mamba~\cite{zhu2024vision} introduced bidirectional state space models by incorporating an additional backward branch to compress visual representations. Hydra~\cite{hwang2024hydra} introduced a quasiseparable matrix mixer to establish a bidirectional extension of Mamba. In this paper, we also experiment with different Mamba variants and provide a detailed comparison of their performance.

\section{Preliminaries}
\subsection{Diffusion Models}
Diffusion models are a class of generative models that have been widely applied in various domains.
The core idea behind diffusion models is to gradually transform a simple noise distribution into a complex data distribution through a sequence of steps. This is achieved through two main processes: the forward process and the reverse process. Next, we introduce the details of DDPM~\cite{ho2020denoising}.

\noindent\textbf{Forward Process.}
The forward process gradually adds noise to the data in a series of $T$ steps. Starting with a data sample $\mathbf{x}_0$, 
Gaussian noise is incrementally added to the sample at each time step $t$. The process is defined by a Markov chain:
\begin{equation}
q(\mathbf{x}_t | \mathbf{x}_{t-1}) = \mathcal{N}(\mathbf{x}_t; \sqrt{1-\beta_t}\mathbf{x}_{t-1}, \beta_t \mathbf{I}),
\end{equation}
where $\beta_t \in (0, 1)$ is a variance schedule that controls the amount of noise added at each step. The forward process can be formulated as: $q(\mathbf{x}_{1:T} | \mathbf{x}_0) = \prod_{t=1}^{T} q(\mathbf{x}_t | \mathbf{x}_{t-1})$.

\noindent\textbf{Reverse Process.}
The reverse process aims to recover the original data sample $\mathbf{x}_0$ from the noisy sample $\mathbf{x}_T$. This is done by learning a reverse Markov chain parameterized by a neural network:
\begin{equation}
p_\theta(\mathbf{x}_{t-1} | \mathbf{x}_t) = \mathcal{N}(\mathbf{x}_{t-1}; \mu_\theta(\mathbf{x}_t, t), \Sigma_\theta(\mathbf{x}_t, t)),
\end{equation}
where $\mu_\theta(\mathbf{x}_t, t)$ and $\Sigma_\theta(\mathbf{x}_t, t)$ are the mean and covariance
at each step $t$. 
The reverse process can be expressed as: $p_\theta(\mathbf{x}_{0:T}) = p(\mathbf{x}_T) \prod_{t=1}^{T} p_\theta(\mathbf{x}_{t-1} | \mathbf{x}_t)$.
The training of the diffusion model involves optimizing the parameters $\theta$ by minimizing the variational bound on the negative log-likelihood of the data, which typically reduces to a mean-squared error (MSE) loss between the true noise and the predicted noise:
\begin{equation}
    \mathcal{L}_{\text{ddpm}} = \left|\left|\bepsilon - \bepsilon_\theta(\sqrt{\bar\alpha_t} \mathbf{x}_0 + \sqrt{1-\bar\alpha_t}\bepsilon, t)\right|\right|^2_2,
\end{equation}
where $\bar\alpha_t$ depends on the noise scheduler. 

It is important to note that our approach employs DDIM~\cite{song2020denoising}, which can be viewed as an extension of DDPM. 
The detailed operational process and advantages of using DDIM will be discussed in subsequent sections.

\begin{figure*}[tbp]
	\centering
	\includegraphics[width=1\linewidth]{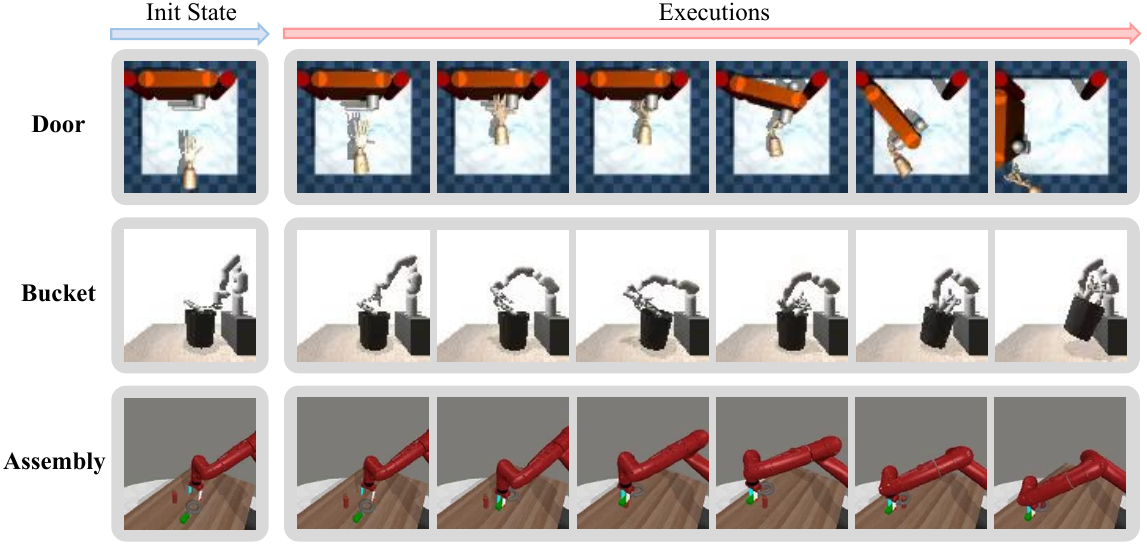}
 \caption{\textbf{Visualization of our manipulation results.} We conduct experiments on three datasets, including Adorit, MetaWorld, and DexArt. Here we illustrate the results in Adroit Door (top), DexArt Bucket (middle), and MetaWorld Assembly (bottom). During the interaction, our proposed Mamba Policy outputs future execution actions until the task is successfully implemented.}
	\label{fig:exp_demo}
\end{figure*}

\section{Our Method: Mamba Policy}

\subsection{Overview}

As illustrated in Fig.~\ref{fig:main}, the Mamba policy is divided into two parts: perception extraction and decision prediction. In the perception stage, we utilize the simple MLP encoder~\cite{ze20243d} to process single-view point clouds for perception extraction. The extracted features, combined with self-state features and time embeddings, are then fed into the X-Mamba UNet, thereby accomplishing decision prediction. Similar to the Diffusion Policy~\cite{chi2023diffusion}, we introduce the concepts of total prediction horizon $T$, observation length $T_o$, and action prediction length $T_a$. At time step $t$, the Mamba policy ingests the most recent $T_o$ steps of observation data as its input and forecasts $T$ steps of action, wherein $T_a$ steps of actions starting from $t$ are as the output.

\subsection{XMamba}
In this study, we develop an innovative denoising network: X-Mamba UNet, with XMamba blocks serving as a key role. We will next delve into the operational process of the XMamba:
 Suppose the input perception feature $c\in \mathbb{R}^{D}$ is processed through the DP3 encoder~\cite{ze20243d} with point cloud and self-state information, the noise input $a\in \mathbb{R}^{d}$ is randomly initialized from Gaussian distribution, where $D$ and $d$ denote the embedding and action dimension, respectively. $c$ and $a$ will be processed through $N$ XMamba block that consists of the fusion stage and refinement stage, and eventually return the prediction.

\noindent\textbf{FiLM Fusion:} To effectively integrate the perception features with the input, we employ the Feature-wise Linear Modulation (FiLM~\cite{perez2018film}) method, where the specific process is presented as follows:
\begin{align}
    x_{in} & = \sigma(\texttt{GN}(\texttt{Conv1D}(a))), \\
    x_{s}, x_{b} &= f(\texttt{GN}(\sigma(c))), \\
    o_1 &= x_{s} * x_{in} + x_b,
\end{align}
where $\sigma$ denotes the mish~\cite{misra2019mish} activation function, $\texttt{GN}$ means the group normalization~\cite{wu2018group}, $f$ denotes the split function which divides the tensor into two parts.

\noindent\textbf{MambaAttn Refinement:} Next, we further refine the features using the Mamba and Attention module. We begin by presenting the definition of the mixer module:
\begin{align}
    & \mathcal{F}(x; \texttt{Mixer}): \nonumber \\
    & \dot{y} = x + \texttt{Drop}(\texttt{Mixer}(x) * \gamma_1), \\
    & y = \dot{y} + \texttt{Drop}(\texttt{MLP}(\dot{y}) * \gamma_2),
\end{align}
where $\texttt{Drop}$ denotes the drop path~\cite{larsson2016fractalnet} strategy, $\gamma_{i}$ denotes the trainable parameter to control the feature scale.
Then, the fusion feature $o_1$ is processed through:
\begin{align}
    o_2 &= \mathcal{F}(o_1; \texttt{Mamba}), \\
    o_3 &= \mathcal{F}(o_2; \texttt{Attention}), 
\end{align}
where we adopt the standard Mamba~\cite{gu2023mamba} and attention~\cite{vaswani2017attention} block. After processing through $N$ XMamba blocks, a simple decoder is utilized to map the feature into our desired shape:
\begin{align}
    a_{out} = \texttt{Conv1D}\big(\sigma(\texttt{GN}(\texttt{Conv1D}(o_3)))\big).
\end{align}

\begin{figure*}[t]
\centering
\includegraphics[width=\linewidth]{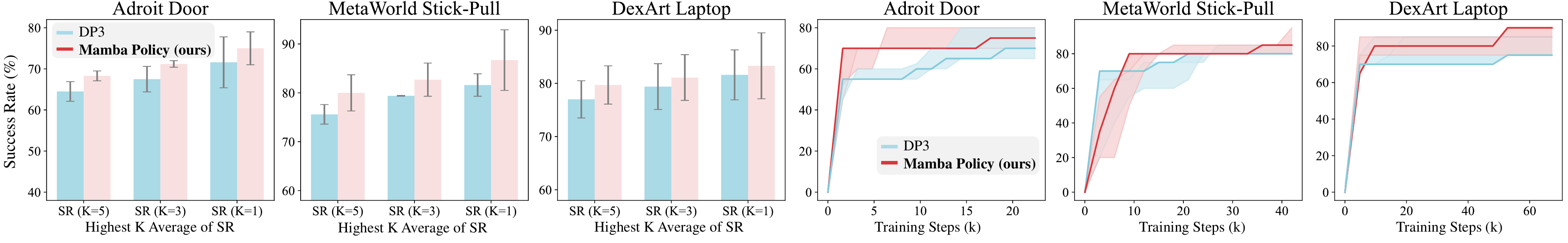}
\caption{\textbf{Visualization of Success Rates and Training Curves.} We visualize the comparisons in terms of different highest $K$ average of success rates, where our proposed Mamba Policy achieves superior results. The stable training curves also demonstrate the effectiveness of our model. }
\label{fig:exp_sr}
\end{figure*}

\begin{table*}[t]
\renewcommand{\arraystretch}{1.1}
\centering
\caption{\textbf{Quantitative comparisons of different baselines in simulation environments.} We compare our Mamba Policy with IBC, BCRNN, 3D Diffusion Policy, and Diffusion Policy in Adroit, MetaWorld and DexArt datasets, in terms of SR$_{5}$. $^\dagger$ denotes our reproduced results for fair comparison. Our proposed Mamba Policy achieves superior results across all domains. }
\label{tab:main_exp}
\resizebox{1\textwidth}{!}{%
\begin{tabular}{l|ccc|ccc|cccc|cc}
\hline\hline

    & \multicolumn{3}{c|}{\textbf{Adroit}} & \multicolumn{3}{c|}{\textbf{MetaWorld}} & \multicolumn{4}{c|}{\textbf{DexArt}} & \\
  Method & \multicolumn{1}{c}{Hammer} & \multicolumn{1}{c}{Door}  & \multicolumn{1}{c|}{Pen} & Assembly & Disassemble & Stick-Push & Laptop & Faucet & Toilet & Bucket & \textbf{Average} \\
\hline
Diffusion Policy~\cite{chi2023diffusion} &	\dd{48}{17} & \dd{50}{5} & \dd{25}{4}	& \dd{15}{1} & \dd{43}{7} & 	\dd{63}{3} & \dd{69}{4} & \dd{23}{8}  & \dd{58}{2} & \dd{46}{1} & $44.0$ \\
BCRNN~\cite{mandlekar2021matters} & \dd{8}{14} & \dd{0}{0} & \dd{8}{1} & \dd{1}{5} & \dd{11}{6} & \dd{0}{0} &  \dd{29}{12} & \dd{26}{2} & \dd{38}{10} & \dd{24}{11} &  $14.5$ \\
IBC~\cite{florence2022implicit} & \dd{0}{0} & \dd{0}{0} & \dd{10}{1} & \dd{18}{9} & \dd{3}{5} & \dd{50}{6} & \dd{1}{1} & \dd{7}{2} & \dd{15}{1} & \dd{0}{0} & $10.4$\\

DP3~\cite{ze20243d}$^{\dagger}$ & \dd{100.0}{0.0} & \dd{64.5}{2.4} & \dd{41.0}{3.3} & \dd{99.6}{0.4} & \ddbf{76.0}{2.4} & \ddbf{100.0}{0.0} & \dd{77.0}{3.5} & \dd{30.0}{2.9} & \dd{74.3}{2.6} & \dd{27.0}{0.8} & $68.9$\\

\hline
\textbf{Mamba Policy (ours)} & \ddbf{100.0}{0} & \ddbf{68.3}{1.2} & \ddbf{41.0}{2.4} & \ddbf{100.0}{0.0} & \ddbf{76.0}{3.7} & \ddbf{100.0}{0.0} & \ddbf{79.6}{3.6} & \ddbf{33.0}{2.1} & \ddbf{76.3}{0.4} & \ddbf{27.0}{0.8} & \ddbfone{70.1}\\

\hline\hline
\end{tabular}}
\vspace{-0.5mm}
\end{table*}

\subsection{Decision Making}
The decision module of our Mamba Policy is based on the conditional diffusion models.
In this section, we will introduce the training and inference process of our Mamba Policy. We simplify the aforementioned XMamba UNet as $\bepsilon_{\theta}$, and we use the DDIM~\cite{song2020denoising} as our diffusion solver.

\noindent\textbf{Training:} The training process starts by randomly drawing action samples $\mathbf{a}_0$ from raw dataset. In the denoising process, the input will be modified into noised action with a random noise $\bepsilon\in\mathcal{N}(0,\mathbf{I})$ for iteration $t$. The denoise network $\bepsilon_{\theta}$ will then predict the noise with input noised and the perception condition. As described in DDIM~\cite{song2020denoising}, the objective is to minimize the KL-divergence between the origin data distribution and the generated sample distribution, we modify the loss function by adding the condition $c$ as follows: 
\begin{equation}
    \mathcal{L}_{\text{ddim}} = \left|\left|\bepsilon - \bepsilon_\theta(\sqrt{\bar\alpha_t} \mathbf{a}_0 + \sqrt{1-\bar\alpha_t}\bepsilon, c, t)\right|\right|^2_2, \label{eq:loss}
\end{equation}
where $\bar\alpha_t$ depends on the noise scheduler.

\noindent\textbf{Inference:}
After training the denoising network $\bepsilon_\theta$, we can now approximate the distribution $p_{\theta}(\mathbf{a}_{t-1}|\mathbf{a}_{t})$ using a non-stochastic approach and iteratively denoise the noised action $\mathbf{a}_{K}$ into the predicted action $a_0$, where $K$ denotes the diffusion steps. Each iterative step is illustrated through:
\begin{align}
    \mathbf{a}_{t-1} &= \sqrt{\bar\alpha_{t-1}}\left( \frac{\mathbf{a}_t - \sqrt{1-\bar\alpha_t}}{\sqrt{\bar\alpha_t}} \bepsilon_\theta(\mathbf{a}_t, c, t) \right) \\
    &+ \sqrt{1-\bar\alpha_{t-1}-\sigma_t^2}\cdot\bepsilon_\theta(\mathbf{a}_t, c, t) + \sigma_t \bepsilon_t, \nonumber
\end{align}
where $\bar\alpha_t, \sigma_t$ are related to the noise scheduler settings, and $\bepsilon_t\in\mathcal{N}(0,\mathbf{I})$. With $K$ steps, we can obtain the final action prediction $\mathbf{a}_0$ to interact in the RL environments.

\section{Experiment}

\noindent\textbf{Datasets.} We conduct our experiments in a large variety of datasets, including 3 domains in Adroit~\cite{rajeswaran2017learning}, 2 domains in MetaWorld~\cite{yu2020meta} with very hard levels, and 4 domains in DexArt~\cite{bao2023dexart} environments. We adopt the same data collection method in DP3~\cite{ze20243d}, where we only generate successful trajectories as expert data.

\noindent\textbf{Baselines.} We choose the 3D diffusion policy (DP3~\cite{ze20243d}), 2D diffusion policy (DP~\cite{chi2023diffusion}), BCRNN~\cite{mandlekar2021matters}, and IBC~\cite{florence2022implicit} as our baselines, where the results are from DP3's origin paper. Since the generation of expert data is subject to randomness, we reproduced DP3 results with marker $\dagger$ by re-generated dataset to ensure fair comparisons.

\noindent\textbf{Experiment Settings.} 
In the primary experiments, we set the prediction horizon to 4, the observation length to 2, and the action prediction length to 3. The X-Mamba UNet is configured with dimensions [128, 256, 512]. Training parameters are consistent with those used in DP3, including a DDIM noise scheduler with a total timestep of 100 and an inference timestep of 10. The AdamW optimizer is utilized with an initial learning rate of 1e-4, and a cosine learning rate scheduler is employed. The model is trained for a total of 3000 epochs with a batch size of 128.

\noindent\textbf{Evaluation Metrics.} We compute the average of the highest 1, 3, 5 success rates and log them as SR$_{1}$, SR$_{3}$, SR$_{5}$, respectively. For each domain, we run 3 seeds (0, 1, 2) and report the mean and standard deviation of the results across three seeds.

\begin{table*}[t]
\renewcommand{\arraystretch}{1.3}
\centering
\caption{\textbf{Efficiency comparisons on Mamba Policy and 3D Diffusion Policy.} We conduct experiments in difficult MetaWorld environments, where our method achieves stronger results with $\mathbf{80\%}$ fewer parameters and $\mathbf{127\%}$ smaller GPU usage. We abbreviate $\texttt{very hard}$ and $\texttt{hard}$ levels as $\texttt{VH}$ and $\texttt{H}$. }
\label{tab:efficiency}
\resizebox{1\linewidth}{!}{%
\begin{tabular}{l|cccc|cccc}
\hline\hline

   \multirow{2}{*}{\centering Method} & \multirow{2}{*}{\centering \# Param (M)} & \multirow{2}{*}{\centering FLOPs (G)} & \multirow{2}{*}{\centering Torch Allocated Memory (MB)} & \multirow{2}{*}{\centering Total GPU Usage (MB)} & \multicolumn{3}{c}{\textbf{Hard Environments in MetaWorld}} \\
   & & & & & \multicolumn{1}{c}{Stick-Pull ($\texttt{VH}$)} & \multicolumn{1}{c}{Disassemble ($\texttt{VH}$)}  & \multicolumn{1}{c}{Pick-Out-Of-Hole ($\texttt{H}$)} \\
\hline
DP3 & \ddone{255.1}  & \ddone{0.30} & \ddone{996.1} & \ddone{8306} & \ddone{81.6} & \ddone{80.0} & \ddone{45.0}\\
{Mamba Policy} & \ddone{47.9} ($\downarrow$~\ddbfone{81.2\%}) & \ddone{0.03} ($\downarrow$~\ddbfone{90.0\%}) & \ddone{137.7} ($\downarrow$~\ddbfone{86.2\%}) & \ddone{3648} ($\downarrow$~\ddbfone{127.7\%}) & \ddbfone{86.7} ($\uparrow$~\ddbfone{5.1\%}) & \ddbfone{81.7} ($\uparrow$~\ddbfone{1.7\%}) & \ddbfone{48.3} ($\uparrow$~\ddbfone{3.3\%})\\
\hline\hline
\end{tabular}}
\end{table*}

\subsection{Comparisons with the State-of-the-Arts}

As shown in Tab.~\ref{tab:main_exp}, we conduct extensive evaluations across a broad spectrum of manipulation datasets to thoroughly assess the effectiveness of our Mamba Policy. The results demonstrate that Mamba Policy significantly surpasses the baseline models, particularly in terms of the SR$_{5}$ metric. For instance, in the Adroit Door domain, our model achieves a notable performance of $68.3$ compared to $64.5$ by DP3, underscoring its superior capabilities. Beyond individual cases, Mamba Policy also consistently exhibits a higher overall average performance across all evaluated domains.
To provide a more detailed understanding of the model's strengths, Fig.~\ref{fig:exp_sr} presents a comparative analysis of the SR$_{1}$, SR$_{3}$, and SR$_{5}$ metrics. These metrics are indicative of the model's performance under varying levels of challenge, where a smaller K in SR$_{K}$ reflects the upper bound of the model’s capabilities, and a larger K highlights its average performance across tasks. Notably, Mamba Policy excels across all metrics, showcasing its robustness and adaptability. Additionally, we have visualized the training process to illustrate the stability and reliability of the model's training, further reinforcing the effectiveness of our approach.

\subsection{Efficiency Analysis}

To assess the computational efficiency of our proposed model, we conduct an efficiency analysis based on the results shown in Tab.~\ref{tab:efficiency}. The analysis reveals that our model not only achieves better performance but does so with significantly lower computational demands. For example, under the Adorit Pen environment, the model parameters are reduced by 81.2\%, from 255.1 to 47.9 M, and memory usage decreased by 90.0\% regarding the number of floating point operations (FLOPs). Additionally, we record the PyTorch allocated GPU memory per loss backward during training and the required GPU usage dropped by 86.2\%. The total GPU usage with our Mamba Policy is 127.7\% smaller that the DP3's. Despite these reductions, our model improves the success rate (K=1) by 5.1\% in Stick-Pull, 1.7\% in Disassemble, and 3.3\% in Pick-Out-Of-Hole. These results underscore the efficiency of our model, demonstrating its ability to deliver superior performance with much lower computational costs.

\begin{figure}[tbp]
	\centering
	\includegraphics[width=.95\linewidth]{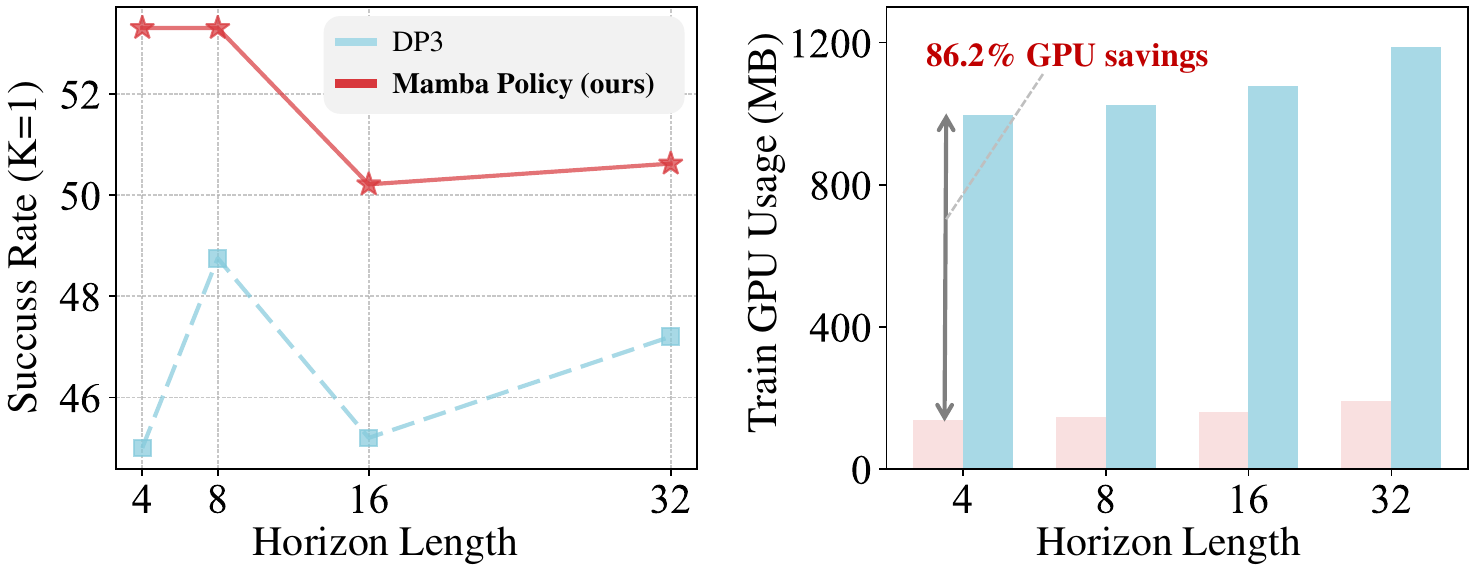}
 \caption{\textbf{Ablation study on different horizon length.} To validate the ability to process longer historical dependencies, we conduct experiments with various horizon lengths and our Mamba Policy achieved robust accuracy and reduced GPU usage compared with DP3, demonstrating the effectiveness and efficiency of our method under long-term scenarios. }
	\label{fig:horizon}
	\vspace{-0.25cm}
\end{figure}

\begin{table}[t]
\renewcommand{\arraystretch}{1.1}
\centering
\caption{\textbf{Ablation study on Mamba Policy with different SSM variants.} We include Mamba, Mamba2, bidirectional SSM, and Hydra for comparisons, where the V1-based and Hydra-based policy achieves good performances.}
\label{tab:ssm}
\resizebox{0.99\columnwidth}{!}{%
\begin{tabular}{l|cccc}
\hline\hline

   \multirow{2}{*}{\centering Method} & \multicolumn{3}{c}{\textbf{Adroit-Pen}} \\
   & \multicolumn{1}{c}{SR$_5$} & \multicolumn{1}{c}{SR$_3$}  & \multicolumn{1}{c}{SR$_1$} \\
\hline

DP3 & \dd{41.0}{3.3} & \dd{42.7}{4.0} & \dd{45.0}{4.0}\\
{Mamba Policy + Mamba-V1~\cite{gu2023mamba}} & \dd{41.0}{2.4} & \ddbf{45.3}{1.9} & \ddbf{53.3}{2.3}\\
{Mamba Policy + Mamba-V2~\cite{dao2024transformers}} & \dd{41.5}{4.5} & \dd{43.1}{4.7} & \dd{45.0}{4.0}\\
{Mamba Policy + Bi-SSM\cite{zhu2024vision}} & \dd{38.9}{4.0} & \dd{43.0}{5.7} & \dd{48.3}{8.4}\\
{Mamba Policy + Hydra~\cite{hwang2024hydra}} & \ddbf{41.8}{6.4} & \dd{45.2}{6.5} & \ddbf{53.3}{6.2}\\
\hline\hline
\end{tabular}}
\vspace{-0.3cm}
\end{table}

\begin{table}[t]
\renewcommand\arraystretch{1.1}
\begin{center}
\caption{\textbf{Ablation study on key components on Adorit Door environment.} The results underscore the necessity of each proposed component.  }
\label{tab:component}
\resizebox{.99\columnwidth}{!}{
\begin{tabular}{cccc|ccc}
\hline
\hline
 Baseline  & Mamba & Attention &MLP & SR$_{5}$ &  SR$_{3}$  &  SR$_{1}$  \\
\hline
\checkmark & & & & \dd{61.3}{2.6}& \dd{65.0}{2.9}& \dd{71.6}{6.2} \\
&\checkmark & & \checkmark&  \dd{66.9}{1.4}& \dd{69.1}{1.7}& \dd{71.6}{2.3}\\
& &\checkmark &\checkmark & \dd{54.8}{6.1}&\dd{56.9}{7.2} &\dd{61.6}{8.4}  \\
& \checkmark& \checkmark  & & \dd{61.0}{0.8}&\dd{64.4}{2.1} &\dd{68.3}{2.3}  \\
& \checkmark& \checkmark & \checkmark & \ddbf{68.3}{1.2} & \ddbf{71.2}{0.8} & \ddbf{75.0}{4.0} \\

\hline
\hline
\end{tabular}}
\end{center}
\vspace{-6mm}
\end{table}

\subsection{Ablation Study}

\noindent\textbf{Ablations on SSM Variants.} As detailed in Tab.~\ref{tab:ssm}, we conduct an ablation study by integrating different structured SSM variants into our model. Incorporating Mamba-V1 results in notable improvements, especially in the SR$_{3}$ and SR$_{1}$ metrics. The Mamba-V2  policy achieves a marginal improvement compared to DP3. In addition, using bidirectional SSM negatively impacts performance.
In contrast, Hydra policy delivers superior results compared to the baselines, securing top-1 SR$_{5}$ and SR$_{1}$ outcomes. However, empirical results indicate that Hydra is very slow to train.
In conclusion, Hydra performs best when considering only the end result, but Mamba-V1 offers better overall performance when the time factor is taken into account.

\noindent\textbf{Ablations on Horizon Lengths.}
In our ablation study on different horizon lengths, we compare the performance of our model against a baseline across various sequence lengths: 4, 8, 16, and 32. Our model consistently outperforms the baseline at each length, demonstrating its robustness across varying temporal horizons. Specifically, for a length of 4, our model achieves a performance score of $53.3$, significantly higher than the baseline's $45.0$. These results suggest that our model's architecture is better suited for handling longer temporal dependencies, yielding more accurate predictions and enhanced robustness in long-horizon scenarios.

\noindent\textbf{Ablations on Key Components.}
We conduct a comprehensive ablation study to evaluate the contributions of the core components in our proposed model, where the baseline only contains the FiLM fusion module. As illustrated in Tab.~\ref{tab:component}, adding the Mamba module alone shows a significant improvement, validating its role in enhancing model robustness. However, with Attention alone or MambaAttn without MLP gains accuracy decreases. The full model including all components achieves the highest success rates, indicating the critical importance of adaptive attention in focusing on relevant features and underscoring the necessity of each proposed method in this work.

\begin{figure}[t!]
	\centering
	\includegraphics[width=.95\linewidth]{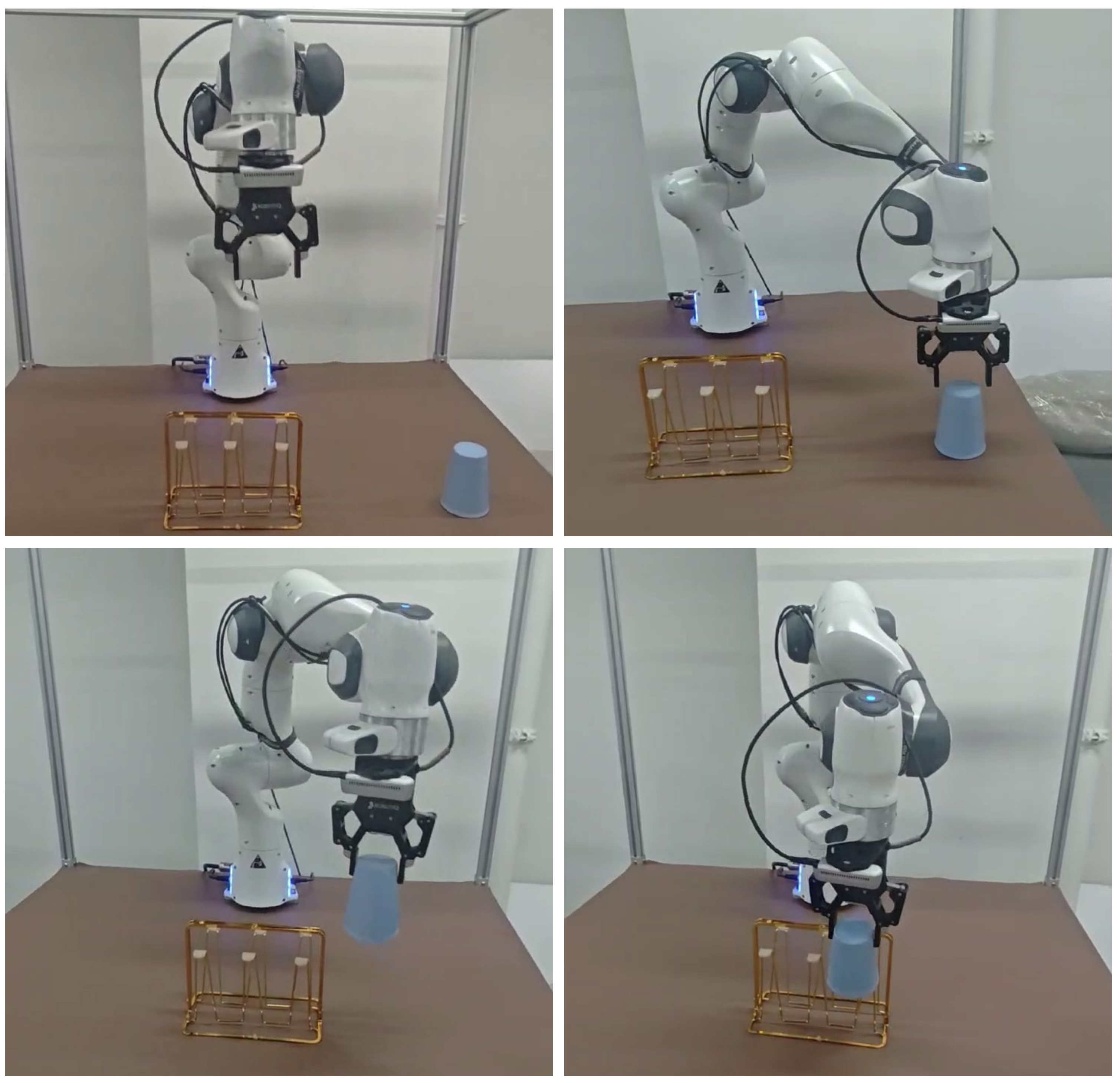}
\caption{\textbf{Real-world experiment.} The robot successfully grasps the cup and places it on the cup holder as part of the 3D manipulation task. The sequence of images demonstrates the robot's precise grasping and placement actions, validating the effectiveness of the proposed Mamba Policy in real-world robotic manipulation tasks.}
	\label{fig:real_world}
	\vspace{-0.25cm}
\end{figure}

\section{Real World Experiment}
\noindent\textbf{Experiment Settings.} In the real-world experiment, we utilize a Franka arm equipped with a gripper, along with a Realsense D455 camera to capture observations during the task. Inference is performed using a 4090 GPU. Our designed task involved the robot grasping a cup and placing it accurately on a cup holder. The robot is required to execute the task with high precision, both in terms of grasping the cup and positioning it correctly on the holder. Data collection is carried out via human teleoperation, where a total of 40 demonstrations are gathered for training the model.

\noindent\textbf{Results.}
As illustrated in Fig.~\ref{fig:real_world}, our Mamba policy demonstrated a high level of precision in executing the grasping operation and successfully placing the cup on the holder. The results validate the effectiveness and practical applicability of the proposed method, showcasing the robot's ability to perform the task with accuracy and reliability in a real-world environment.

\section{Conclusion}

In this paper, we introduced the Mamba Policy, a lightweight yet efficient model tailored for 3D manipulation tasks. By significantly reducing the parameter count by over 80\% compared to 3D diffusion policy, the Mamba Policy maintains strong performance while being more suitable for deployment on resource-constrained devices. Central to this approach is the XMamba Block, which effectively combines input data with conditional features through the integration of Mamba and Attention mechanisms.
Our extensive experiments on the Adroit, Dexart, and MetaWorld datasets demonstrate the Mamba Policy's superior performance and reduced computational requirements. Additionally, we showed that the Mamba Policy exhibits enhanced robustness in long-horizon scenarios, making it a compelling choice for various real-world applications.
This work sets the stage for future research in developing efficient models for 3D manipulation, particularly in environments with limited computational resources. Real-world experiments are also conducted to validate its effectiveness further. Our open-source project page can be found at \href{https://andycao1125.github.io/mamba_policy/}{this link}.

{\small
\bibliographystyle{IEEEtran}
\bibliography{IEEEexample}
}

\end{document}